\documentclass[letterpaper, 10 pt, conference]{ieeeconf}  

\IEEEoverridecommandlockouts                              

\makeatletter
\let\NAT@parse\undefined
\makeatother
\usepackage[numbers]{natbib}

\usepackage[pdftex]{graphicx}
\usepackage{amsmath} 
\usepackage{amssymb}  
\usepackage{subfigure}
\usepackage{multirow}
\usepackage{array,booktabs}
\usepackage{diagbox}
\usepackage{balance}
\usepackage[ruled]{algorithm}
\usepackage{algpseudocode}

\usepackage{irap_SIunits}
\usepackage{irap_acronyms}
\usepackage{irap_math}
\usepackage{irap_misc}

\usepackage{soul,color}
\usepackage{lipsum}

\usepackage{hyperref}
\hypersetup{
    colorlinks=true,
    linkcolor=blue,
    filecolor=magenta,
    urlcolor=cyan,
}

\usepackage{xcolor}

\usepackage{caption}

\usepackage{ulem}
\usepackage{tikz} 

\title{\LARGE \bf
Balanced Depth Completion between Dense Depth Inference\\and Sparse Range Measurements via KISS-GP
}

\author{Sungho Yoon${}^{1}$ and Ayoung Kim${}^{2*}$
\thanks{$^{1}$S. Yoon is with the Robotics Program,
        KAIST, Daejeon, S. Korea
        {\tt\small sungho.yoon@kaist.ac.kr}}%
\thanks{$^{2}$A. Kim is with the Department of Civil and Environmental Engineering,
        KAIST, Daejeon, S. Korea
        {\tt\small ayoungk@kaist.ac.kr}}%
\thanks{This work was fully supported by [Localization in changing city] project funded by Naver Labs Corporation.}
}

\begin{document}

\maketitle
\thispagestyle{empty}
\pagestyle{empty}

\begin{abstract}

Estimating a dense and accurate depth map is the key requirement for autonomous driving and robotics. Recent advances in deep learning have allowed depth estimation in full resolution from a single image. Despite this impressive result, many deep-learning-based \acf{MDE} algorithms have failed to keep their accuracy yielding a meter-level estimation error. In many robotics applications, accurate but sparse measurements are readily available from \ac{LiDAR}. Although they are highly accurate, the sparsity limits full resolution depth map reconstruction. Targeting the problem of dense and accurate depth map recovery, this paper introduces the fusion of these two modalities as a  \acf{DC} problem by dividing the role of depth inference and depth regression. Utilizing the state-of-the-art \ac{MDE} and our \acf{GP} based depth-regression method, we propose a general solution that can flexibly work with various MDE modules by enhancing its depth with sparse range measurements. To overcome the major limitation of \ac{GP}, we adopt \ac{KISS}-\ac{GP} and mitigate the computational complexity from $O(N^3)$ to $O(N)$. Our experiments demonstrate that the accuracy and robustness of our method outperform state-of-the-art unsupervised methods for sparse and biased measurements.

\end{abstract}

\section{Introduction}
\label{sec:intro}

Depth sensing is one of the most essential features for autonomous driving and mobile robots so as to avoid obstacles and to build localization and mapping. Toward this objective, industries seek efficient methods to build a dense depth-sensing system with low prices.

Toward this objective, \acf{MDE} is used to predict a dense depth map only with a single image. It has been actively studied in computer vision and robotics area. Recent researches have found that the depth scale can be learned and predicted even in the single-camera setup. Using such property, the recent \acf{SLAM} studies have utilized \ac{MDE} to reduce the uncertainty of the depth initialization, improving its performance \cite{Tateno17, Loo18, Yang2018}. However, it tends to have some limitations in its accuracy because it needs to learn many types of monocular cues and to reason the dense depth map based on the overall features. Moreover, \ac{MDE} does not use any geometric relationship such as triangulation, which signifies that it is a fundamental ill-posed problem.

\begin{figure}[!t]
	\def\width{1\columnwidth}%
	\centering
	\includegraphics[width=0.9\width]{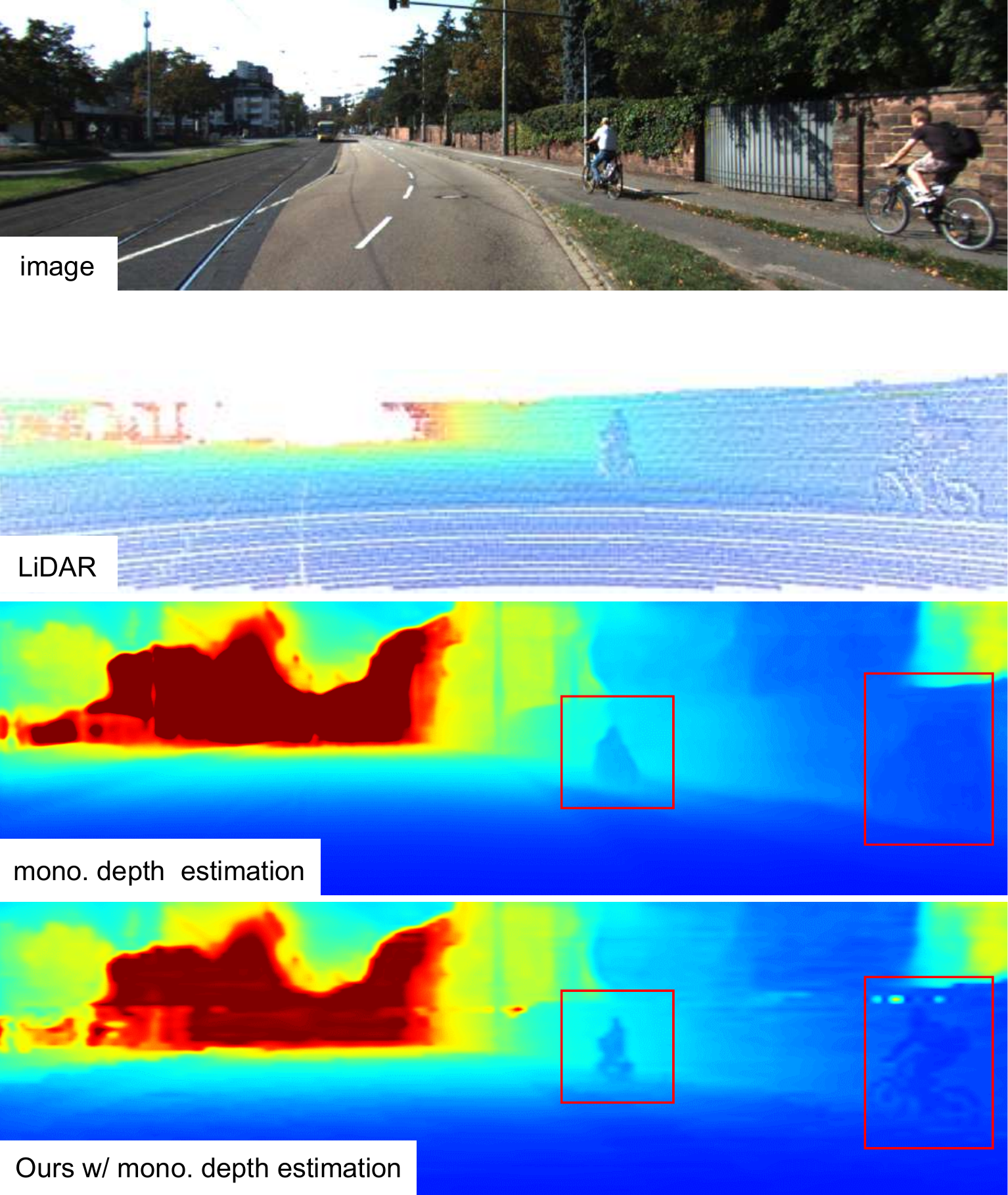}
	\caption{Depth completion in the KITTI dataset. The dot size is inflated for visualization (the second row). The third-row image shows that the shapes of men and bicycles are blurry as a result of the monocular depth estimation using an image. Our method improves the exact shapes using LiDAR and offers robustness on sparse or biased depth measurements.}
	\label{fig:main}
	\vspace{-5mm}
\end{figure}

Another topic, \acf{DC}, is used to boost depth inference performance using both an image and depth measurements. While accurate dense mapping has not been achieved only using the 3D feature points of SLAM or the sparse-point cloud of a lightweight LiDAR, \ac{DC} accompanied by an image has presented a more powerful performance. Similar to \ac{MDE}, this topic has some challenging problems for the three reasons. First, it has difficulties handling multiple sensing modalities such as image and depth measurements. Second, the 3D measurements are highly sparse and irregularly spaced depending on the environment and sensor characteristics.  Third, finding a feasible dataset with well-labeled ground truth is still lacking.

\begin{figure*}[!t]
	\def\width{0.8\textwidth}%
	\centering
	\includegraphics[width=\width]{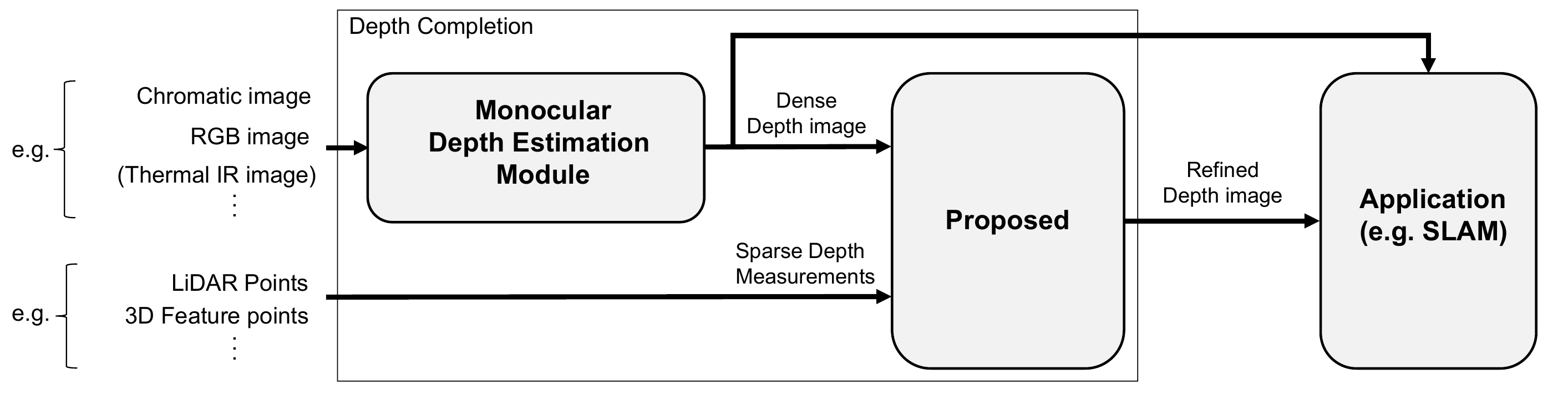}
	\caption{System diagram. The proposed method utilizes a state-of-the-art depth prediction module specifically designed for each imaging modality. Then together with sparse depth measurements such as LiDAR, a depth regression is performed to make a refined depth image using the \acf{GP}.  }
	\label{fig:flow_chart}
	\vspace{-5mm}
\end{figure*}

In this work, we propose a back-end module wherein the result of \ac{MDE} can be directly improved using sparse depth measurements, as shown at \figref{fig:main}.

\begin{itemize}
	\item We propose a general solution for the \ac{DC} problem, which can be used with various state-of-the-art \ac{MDE} modules by means of monochrome, thermal, RGB, or multiple cameras.
	\item The proposed method is especially robust in the cases where the sparsity or bias of range measurements can vary or become unexpectedly deformed. Compared to the deep-learning-based \ac{DC} methods that work appropriately only in their trained situation, our GP-based method maintains stable performance even in changing sensor conditions.
	\item Our evaluation method reveals the effect of changing sparsity or bias of range measurements, which we may have overlooked in the \ac{DC} problem.
\end{itemize}

\section{Literature Review}
\label{sec:review}

\textbf{Monocular Depth Estimation (MDE): }\ac{MDE} aims to predict dense depth from a single image. The classical methods utilize meaningful cues on an image such as perspective projection, texture gradients, shadow, focus/defocus, or relative size~\cite{Subbarao1994, Hoiem2007, Ladicky2014}. Recently, the performance of depth estimation has been significantly improved by exploiting deep neural networks. \citeauthor{Fu2018}~\cite{Fu2018} proposed a supervised depth estimation using a spacing-increasing discretization (SID) strategy and an ordinal regression loss. \citeauthor{Godard2017}~\cite{Godard2017} trained monocular depths in a self-supervised manner using a left-right depth consistency term with a stereo camera. They also proposed another self-supervised method, called Monodepth2, using minimum reprojection loss and a multi-scale sampling method~\cite{Godard2018}. \citeauthor{Liu2019}~\cite{Liu2019} estimated not only its depths but also its uncertainty. \citeauthor{Yin2019}~\cite{Yin2019} improved their depth-prediction accuracy by enforcing the geometric constraints of virtual normal direction to their loss function. \citeauthor{Kuznietsov2017}~\cite{Kuznietsov2017} introduced a training method for their semi-supervised approach using sparse \ac{LiDAR} measurements for a supervised depth cue and using stereo image pairs for an unsupervised training cue. As another semi-supervised approach, \citeauthor{Ramirez2018}~\cite{Ramirez2018} showed that learning depth with semantic labels improved the estimation performance. \citeauthor{Chen2019}~\cite{Chen2019} presented a neural network inferring geometric depth and semantic segments at the same time as their content consistency check.

\textbf{Depth Completion (DC): }\ac{DC} utilizes sparse range measurements to predict an image with a denser depth. \citeauthor{Ma2016}~\cite{Ma2016} approached the \ac{DC} problem within the compressive sensing technique. By minimizing the number of edge candidates given sparse depth measurements, they recovered a 3D map within its error bounds. As more recent works have started using deep learning, they have introduced a supervised auto-encoder architecture that uses an RGB image and sparse depth measurements as inputs~\cite{Ma2017}. DeepLiDAR~\cite{Qiu2018} presented a method using deep-surface normals as an intermediate representation toward dense depth in an outdoor scene. Their network also predicted a confidence mask to select the reliable raw data of LiDAR. \citeauthor{Cheng2018}~\cite{Cheng2018} proposed a convolutional spatial propagation network (CSPN) that learned the affinity matrix for depth estimation. The CSPN has been effective in recovering structural details of the depth map. They~\cite{Cheng2019} further developed CSPN++, which improved their effectiveness and efficiency through context and resource-aware CSPN. The CSPN++ learned two hyperparameters to select adequate convolutional kernel sizes and the number of iterations from the predefined architecture. \citeauthor{Gansbeke2019}~\cite{Gansbeke2019} implemented a global and local depth map, and their depth maps were merged using each confidence map for the final depth.

The recent studies have reported limitations in supervised methods in terms of generalizability over various environments. As an alternative, unsupervised or self-supervised methods were also actively studied in \ac{DC}. \citeauthor{Ma2018}~\cite{Ma2018} proposed a self-supervised method that adds a photometric loss and a second-order smoothness prior. As another unsupervised work, \citeauthor{Yang2019_DDP}~\cite{Yang2019_DDP} exploited a conditional prior network (CPN) that estimated a probability of depths at each pixel. They trained the CPN in a virtual dataset that is similar to and has been tested in the KITTI depth completion dataset. However, the CPN failed in generalizing to another indoor dataset, NYU-Depth-V2.

In contrast to these previous unsupervised works, our method is a general \ac{DC} solution with improved stability. Most of the \ac{DC} research highly coupled the RGB image and depth measurements by concatenation at the early steps of their architecture. The existing approaches may lose flexibility when another imaging modality becomes additionally available. Furthermore, under sparse or biased range measurements, the robustness of performance is broken easily, which critically deteriorates safety and stability. Differing from these previous studies and overcoming limitations, we propose a loosely coupled method that divides a \ac{DC} problem to \ac{MDE} and depth regression. Our method overcomes such sparsity or bias issues of range measurements by propagating the precise depths within a limited length using \ac{GP}.

\section{Proposed Method}
\label{sec:method}

This paper reports a solution to combine dense depth estimation from deep learning with sparse-but-accurate depth measurement using \ac{GP} as a regression back-end. The overall system architecture of our method is illustrated in \figref{fig:flow_chart}. We also discuss a depth propagation strategy in the case of measurements like LiDAR that are regularly distanced.

\subsection{Depth Measurements}
\subsubsection{Deep-Learning-Based Monocular Depth Estimation}

Recently introduced \ac{MDE} predicts a depth map of a single image through the \ac{CNN} architecture. Estimating the depth map from a single image is truly an advantage; however, depth accuracy is still very limited compared to the existing model-based triangulation or direct measurement. In our pipeline, we adopted monodepth2~\cite{Godard2018} to obtain depth estimation in a full-size image. For example, the depth estimation \ac{RMSE} of the monodepth2 ranges from 3 to 4 meters when tested over the KITTI odometry and \ac{DC} dataset.

\subsubsection{Sparse Depth Measurement}

Unlike the depth inference of \ac{MDE}, 3D depth can be measured directly and indirectly. An active depth sensor like \ac{LiDAR} is a direct method that utilizes illuminating laser light and measures reflected light. As an indirect method with a monocular camera, depth can be measured through triangulation between corresponding feature points and it is refined by optimization like \acf{BA}. However, both direct and indirect methods provide sparse measurements. The densest \ac{LiDAR} sensor is 128 rays which is still quite sparse when projected on an image plane with a significant cost for a sensor.

\subsection{Depth Integration via KISS-GP}
\label{sec:gp}

\subsubsection{Problem Formulation using Exact GP}

To create a more precise map than \ac{MDE} and a denser map than sparse depth measurement, two different properties should be integrated effectively. The proposed method using \ac{GP} estimates the posterior depth map $\mathbf{Y_*}$ as a mean function $\mu_*$ and its variance $\Sigma_*$. We assume the joint probability of Gaussian random variables between test data $\mathbf{X_*}=\lbrace \mathbf{x_{1:M}}\rbrace$ and training data $X= \lbrace \mathbf{x_{1:N}}\rbrace $, where $\mathbf{x_i}$ is an i-th pixel coordinate on an image $\mathbf{x_i} = \mathbf{(u_i,v_i)}$ and $\mathbf{M}$ and $\mathbf{N}$ are numbers of pixels. In our problem, we set the size $\mathbf{M}$ as the number of pixels in a refined depth map and the size $\mathbf{N}$ as the number of pixels in both imaging depth map and ranging measurements. The size $\mathbf{M}$ of $\mathbf{X_*}$ can be larger than the size $\mathbf{N}$. This shows that we can enlarge the output depth map without linear extrapolation within \ac{GP}. The predictive mean and covariance of $\mathbf{Y_*}$ for a test input $\mathbf{X_*}$ is obtained within a conditional Gaussian distribution $p(Y_*|\mathbf{X}_*,X,Y) = \mathcal{N}(\mu_*, \Sigma_*)$:
\begin{eqnarray}
  \mu_* &=& K(\mathbf{X}_*,X)^\top(K+\sigma_n^2I)^{-1}\mathbf{Y}\\
  \Sigma_* &=& K(\mathbf{X_*},\mathbf{X_*})-K(\mathbf{X_*},X)^\top(K+\sigma_n^2I)^{-1}K(\mathbf{X_*},X)\nonumber.
  \label{eq:gp_mean_cov}
\end{eqnarray}
In the equation, $K$ represents a kernel function $K(X,X)$, and $\sigma_n$ is the observation noise of $Y$.

\subsubsection{Training Set}

For targeting a full-resolution depth image of $m \times n$, we used incoming deep learning-based depth estimation ($X_{DL}= \lbrace \mathbf{x_{1:mn}}\rbrace$) and the $p$ number of measured sparse-but-accurate depth ($X_{Meas}= \lbrace \mathbf{x_{1:p}}\rbrace$) for training data. This training set of $X = \lbrace X_{DL}, X_{Meas}\rbrace$ is the grid coordinates of depth image $(u,v)_{i=1:mn+p}$, and $y_{i=1:mn+p}$ is its depth values $D(u,v)_{i=1:mn+p}$.

\begin{algorithm}[t]
  \caption{{\small Depth Completion with KISS-GP}}
  \label{alg:dc_gp}
  {\small
  \begin{algorithmic}[1]
  \State {\bf Input: }Dense depth image $I_{m \times n}$, Sparse measurement image $J_{m \times n}$, Observation noise $\sigma_{DL}$ and $\sigma_{meas}$, and length-scale $l$
  \State Set of training inputs $X=\lbrace(u_i,v_i)_{1:mn + p}\rbrace$, \\
  Set of training outputs $Y=\lbrace(d_i)_{1:mn + p}\rbrace$, \\
  Set of testing inputs $X_{*}$$ \gets $ \Call {MatToArray}{$D_{in}= I_{m' \times n'}$}
  \State {$K_{X_{*}X} \gets $ \Call {Compute-KernelMat} {$X_{*}, X, l$}}
  \State {$Y_{*} \gets $ \Call {Compute GP-Mean}{$K_{X_{*}X}\lbrace K_{SKI} + \sigma_{mn+p} I \rbrace^{-1} Y$}}
  \State {$D_{out} \gets $ \Call {ArrayToMat}{$Y_{*}$}}
  \State {\bf Output: } Refined depth image $D_{out}$

  \end{algorithmic}
}
\end{algorithm}

\subsubsection{KISS-GP Implementation}

The main challenge in using exact \ac{GP} is its high computation $O(n^3)$ and storage cost $O(n^2)$, along with the fact that it requires calculating a large inverse matrix of kernels. As shown in equation (1), we have been limited to handle large data such as a high-resolution image. To alleviate this issue, we adapted \ac{KISS}-\ac{GP} \cite{Wilson2015} that exploits kernel approximation strategy. By combining an inducing point method, such as a Subset of Regression (SoR) with kernel interpolation, the size of the kernel is approximated into a small size:
\begin{eqnarray}
  \begin{aligned}
K(X,X) & \stackrel{\mathrm{SoR}}{\approx} K(X, U) K(U, U)^{-1} K(U, X)\\
&\stackrel{mat. intrpl}{\approx} W K(U, U) K(U, U)^{-1} K(U, U) W^{\top}\\
&=W K(U, U) W^{\top}=K_{\mathrm{SKI}}.
\end{aligned}
  \label{eq:K_(SKI)}
\end{eqnarray}
The inducing point set $U$ represents the effectively summarized data among the original input data by SoR, and the interpolation weights matrix $W$ contains the relative distances from original input data $X$ and the inducing point $U$. Since the $W$ is extremely sparse, $K_{\mathrm{SKI}}$ approximates the original GP kernel function $K$ with $O(n+m^{2})$ computation and storage. In addition to \ac{SKI}, \ac{KISS}-\ac{GP} in 2D inputs uses a Kronecker structure, and it finally reduces computation $O(2m^{1+1/2})$ and storage $O(n+2m)$.
In practice, we exploited a GPyTorch implementation \cite{Gardner2018} to enhance computational speed further. The GPyTorch is a modern GP toolbox using PyTorch \cite{Paszke2019}, which reduces computational complexity using \ac{BBMM} with GPU acceleration. The \ac{BBMM} uses a modified conjugated gradients algorithm and enables the effective use of GPU hardware \cite{Gardner2018}.

\begin{figure}[!t]
	\def\width{0.9\columnwidth}%
	\centering
	\subfigure[RGB image]{
		\includegraphics[width=\width]{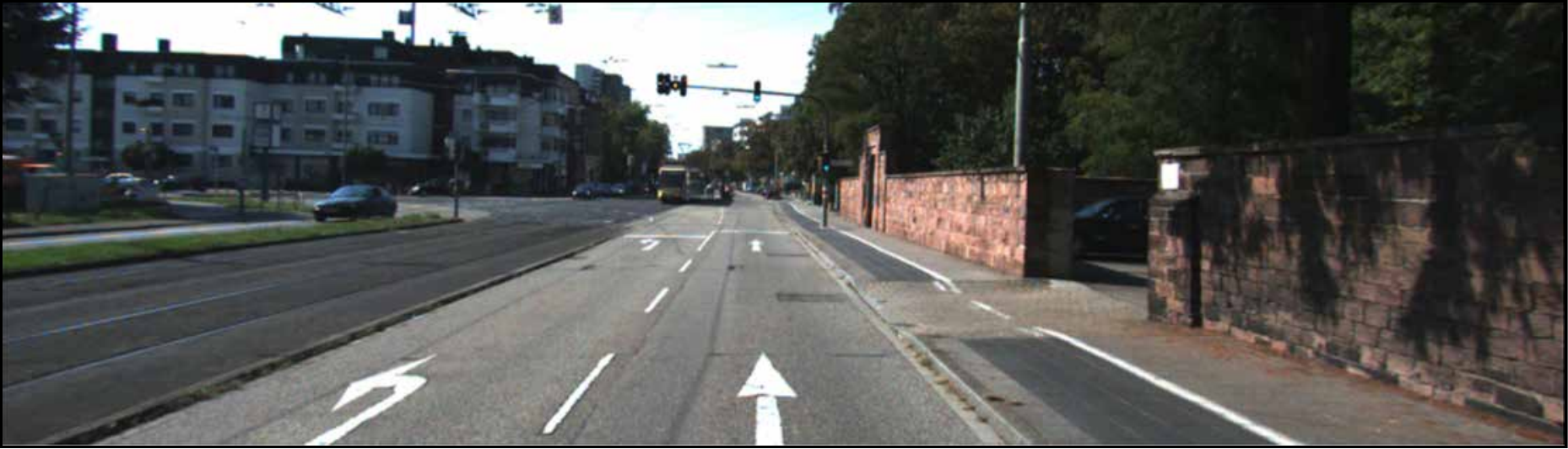}
		\label{fig:rgb}
	}\\
	\subfigure[full scan]{
		\includegraphics[width=\width]{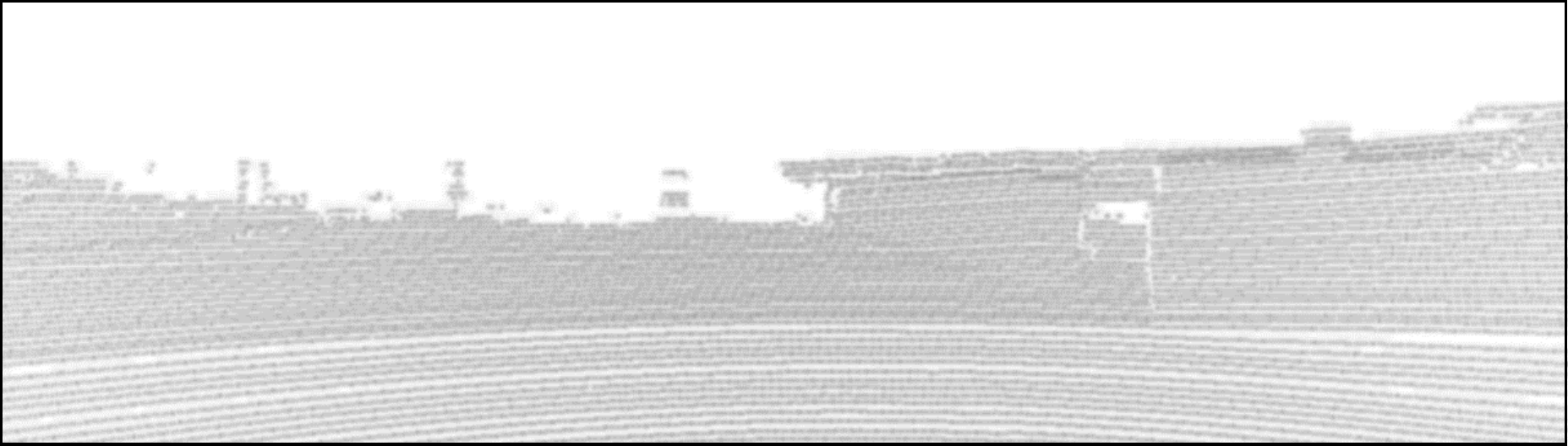}
		\label{fig:full}
	}
	\subfigure[uniformly sampled scan]{
		\includegraphics[width=\width]{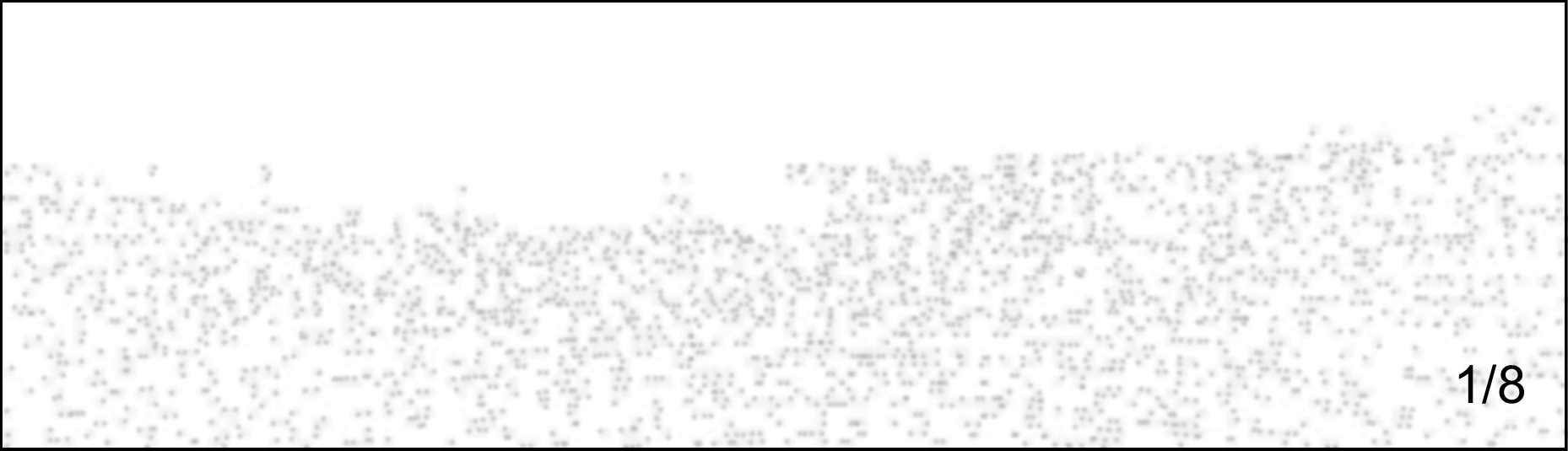}
		\label{fig:uniform}
	}
	\subfigure[horizontally sampled scan]{
		\includegraphics[width=\width]{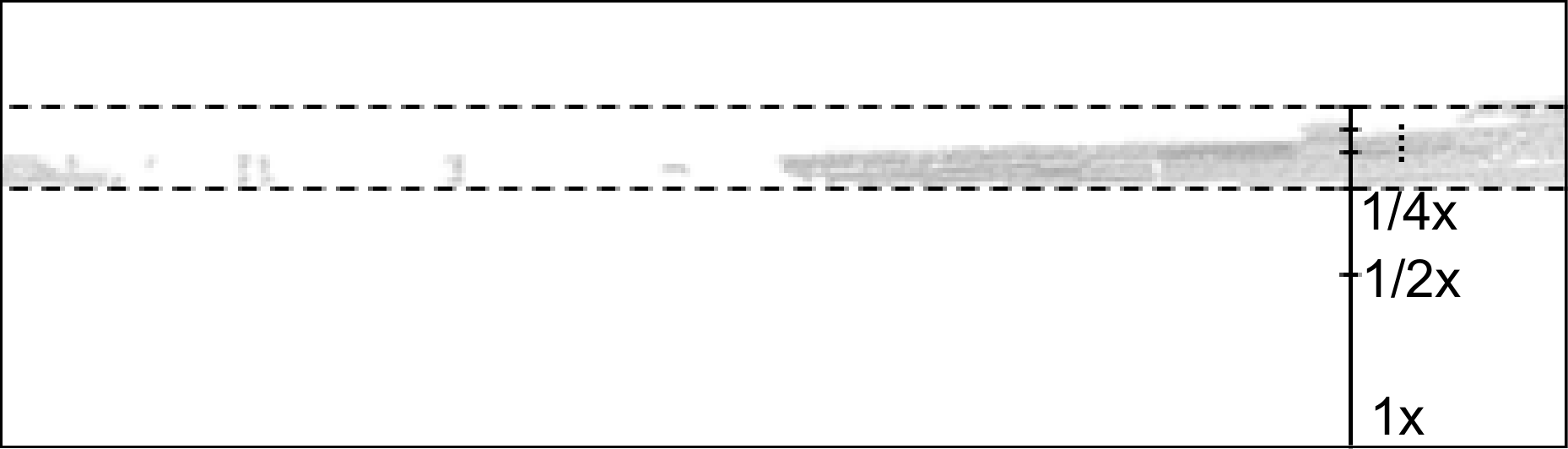}
		\label{fig:horizontal}
	}
	\subfigure[vertically sampled scan]{
		\includegraphics[width=\width]{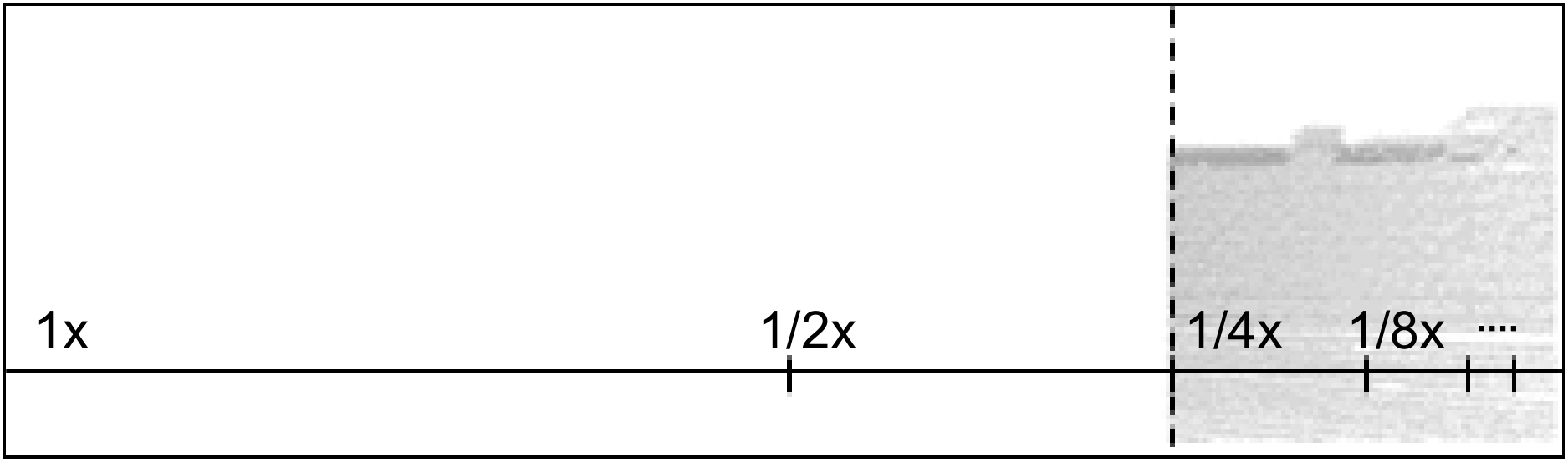}
		\label{fig:vertical}
	}
	\caption{Three different sampling strategies for the evaluation of sparsity and biased measurements. \subref{fig:rgb} and \subref{fig:full} The original RGB image and \ac{LiDAR} raw data. \subref{fig:uniform} The uniform sampling. \subref{fig:horizontal} and \subref{fig:vertical} The full LiDAR data is subsampled into eight levels of biased measurements. The performance of depth extrapolation can be evaluated with horizontally and vertically sampled data. } \label{fig:sampling_methods_KITTI}
  \vspace{-4mm}
\end{figure}

\subsubsection{Kernel and Hyperparameters}

We have selected the Matern kernel for our method, which is a stationary kernel to utilize grid interpolation in KISS-GP. Among stationary kernels like squared exponential, the Matern kernel is more beneficial in incorporating potentially abrupt change near the edges. In terms of hyperparameters, the observation noises and length scale should be defined.  The depth observation noises from two different modalities are different in our case. While the sparse measurements are precise (i.e., low noise), the dense depth image would have higher noise. Since the different sensors have dissimilar accuracy, the observation noise should be uniquely defined depending on each depth model. We used two observation noise terms from different sensor modalities, namely $\sigma_{DL}$ and $\sigma_{Meas}$.

Together with these observation noises, we needed to determine the length scale. Length scale $l$ defines the amount of depth propagation between each pixel. Although it is related to the measurements' sparsity, it is correlated with the noise factor because it is better for the length scale to be large when the observation noise of a dense depth image is low enough. In the \ac{LiDAR} case, in which measurements are almost regularly distributed, the length scale can be easily assigned with the distance between each measurement point. In this paper, we used $\sigma_{DL}=0.05$, $\sigma_{Meas}=0.001$ and $l=1.5$ for KITTI.

\section{Experiments and Results}
\label{sec:results}

In this section, we present a quantitative and qualitative evaluation of the proposed method when tested on both indoor and outdoor. The result is also available from \url{https://youtu.be/x8n0lvjvorg}.


\begin{table}[!b]
	\centering
	\resizebox{\linewidth}{!}{
	\begin{tabular}{c|l|cc}
		\hline
		Metric     & Details                    			& KITTI       & NYU-Depth-V2\\ \hline\hline
		RMSE 		   & root mean squared error					& \checkmark & \checkmark \\ \hline
		MAE  		   & mean absolute error							& \checkmark &  \\
		InvRMSE    & inversed root mean squared error & \checkmark &  \\
		InvMAE     & inversed mean absolute error     & \checkmark &  \\
		REL	 		   & mean absolute relative error     &            & \checkmark \\
		$\delta_i$ & \% of pixel count where          &            & \checkmark \\
		           &  its relative error is within $1.25^i$\\
		\hline
	\end{tabular}
	}
	\caption{Evaluation metrics}
	\label{tab:metrics}
\end{table}

\subsection{Datasets}

For the evaluation of our method, we used two different datasets on outdoor and indoor: KITTI depth completion \cite{Uhrig2017} and NYU-Depth-V2 \cite{Silberman2012}.

In the KITTI depth completion dataset, we resized the original image to one third. When projecting \ac{LiDAR} onto an image plane, occlusion in \ac{LiDAR} may cause duplicated depth measurement for a pixel. Existing \ac{DC} studies resolved this problem by designing a \ac{LiDAR} confidence map \cite{Qiu2018, Gansbeke2019, Eldesokey2018}. By learning the \ac{LiDAR}'s confidence map from a training dataset, they filtered out such unreliable duplicated depth data. Instead of using the trained mask, we alleviated such occlusion effects by selecting the closest distance in a downsampling procedure. For sparsity and bias of depth measurements, we evaluated the performance over three sampling strategies, as shown in \figref{fig:sampling_methods_KITTI}. In the NYU-Depth-V2 dataset, we resized the original images to be half their size and center-cropped them to be $304 \times 228$, performed in previous works~\cite{Ma2017, Qiu2018}. In the experiment, we used a PC with an Intel Core i7-6700 CPU with 3.4GHz, 64GB of RAM, as well as an NVidia GeForce RTX 2080Ti GPU.

As a result of the evaluation, the proposed method directly improves the frond-end \ac{MDE} when additional sparse measurements are given. The accuracy of the proposed method outperforms a state-of-the-art method that uses unsupervised learning in terms of sparsity and bias of \ac{LiDAR} measurements, and it is also comparable with the full \ac{LiDAR} scan case. In the following section, we present experimental results at each dataset. \tabref{tab:metrics} lists the error metrics used in the evaluation. The \ac{RMSE} is the main evaluation metric in both datasets.


\begin{figure*}[!t]
	\def\width{1\textwidth}%
	\centering
	\includegraphics[width=\width]{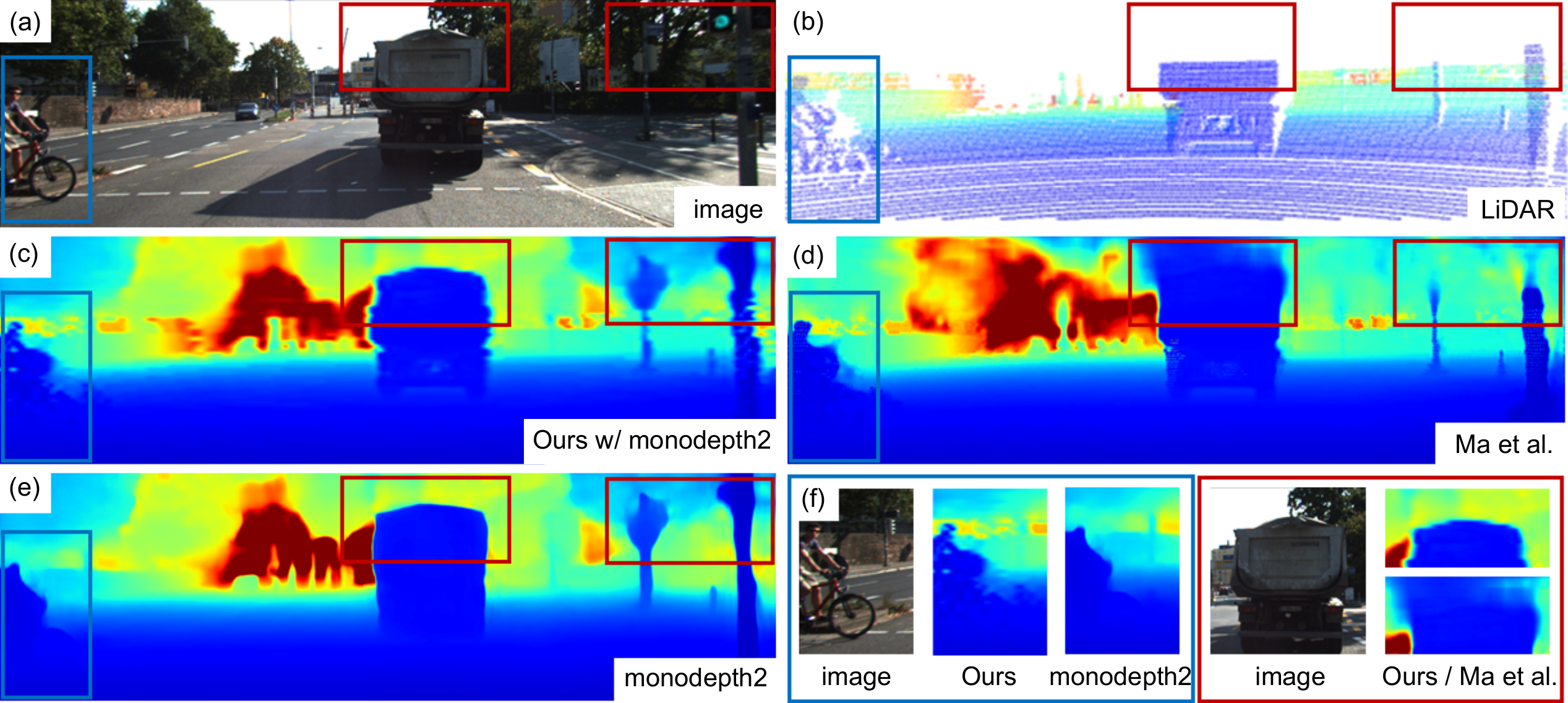}
	\caption{Qualitative comparison to monodepth2~\cite{Godard2018} and \citeauthor{Ma2018}~\cite{Ma2018} over the KITTI dataset. In all sample figures, the warmer color denotes farther distance. (a) and (b) The RGB image and \ac{LiDAR} raw data. Depth map from (c) the proposed method, (d)  \citeauthor{Ma2018}~\cite{Ma2018}, and (e) monodepth2~\cite{Godard2018} are shown respectively. The blue box represents our improvement from monodepth2. The red box illustrates that our method keeps the shape of a truck better than  \citeauthor{Ma2018}~\cite{Ma2018} in the region without \ac{LiDAR} data.}
	\label{fig:kitti_qual}
\end{figure*}

\begin{figure*}[!t]
	\def\width{1\textwidth}%
	\centering
	\includegraphics[width=\width]{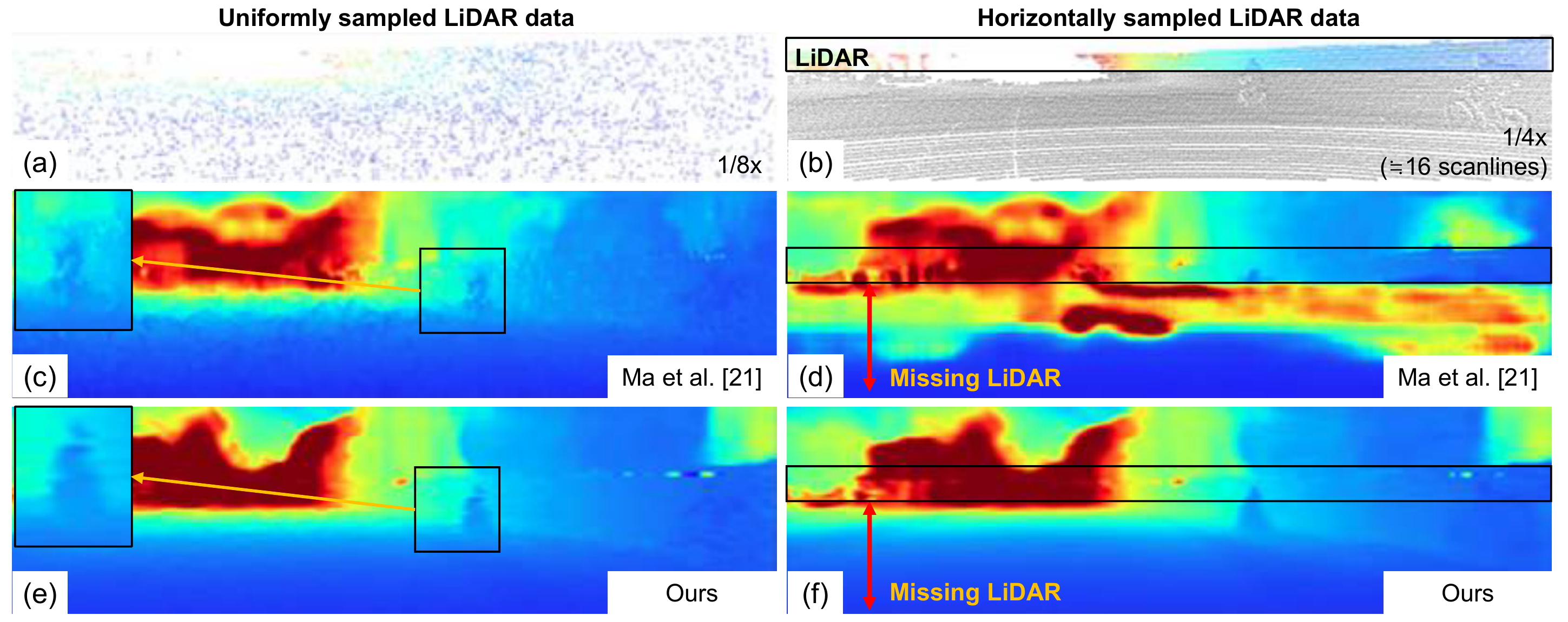}
	\caption{Qualitative comparison to \citeauthor{Ma2018}~\cite{Ma2018} given the sparse and bias LiDAR measurements in the KITTI dataset. (a) and (b) Uniformly sampled \ac{LiDAR} data and horizontally sampled \ac{LiDAR} data. The depth map from (c) and (d) result from \citeauthor{Ma2018}~\cite{Ma2018}. And the depth maps from (e) and (f) result from the proposed method. The results show that the proposed method is more robust on the sparsity or bias change of range measurements than \citeauthor{Ma2018}~\cite{Ma2018}.}
	\label{fig:kitti_qual2}
\end{figure*}

\begin{figure*}[!t]
	\centering
  \begin{minipage}{0.48\textwidth}
    \subfigure[RMSE comparison to \cite{Ma2018} against \ac{LiDAR} points' sparsity and bias]{
	    \includegraphics[trim=4.3cm 15cm 3cm 9.5cm clip, width=\textwidth]{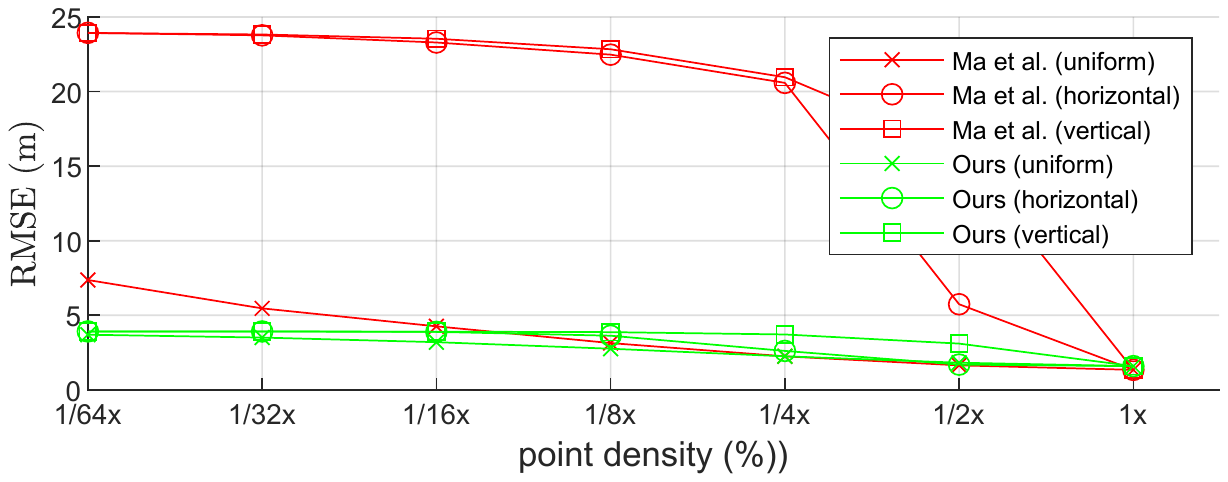} 
	    \label{fig:qual_results_KITTI}
    }
  \end{minipage}%
  \begin{minipage}{0.48\textwidth}
    \subfigure[RMSE metric result (mm)]{
	    \label{tab:results_KITTI}
	    \resizebox{1.0\columnwidth}{1.5cm}{%
	    \begin{tabular}{c|c|c|c|c|c|c}
	    \hline
	      \begin{tabular}[c]{@{}c@{}}point density\end{tabular} &
	      \begin{tabular}[c]{@{}c@{}}Ma et al.\cite{Ma2018} \\ (uniform)\end{tabular} &
	      \begin{tabular}[c]{@{}c@{}}Ma et al.\cite{Ma2018}\\ (horizontal)\end{tabular} &
	      \begin{tabular}[c]{@{}c@{}}Ma et al.\cite{Ma2018}\\ (vertical)\end{tabular} &
	      \begin{tabular}[c]{@{}c@{}}Ours w/ \cite{Godard2018}\\ (uniform)\end{tabular} &
	      \begin{tabular}[c]{@{}c@{}}Ours w/ \cite{Godard2018}\\ (horizontal)\end{tabular} &
	      \begin{tabular}[c]{@{}c@{}}Ours w/ \cite{Godard2018}\\ (vertical)\end{tabular} \\ \hline
	    \begin{tabular}[c]{@{}c@{}}0 (\ac{MDE})\end{tabular}      & 24092.25 & 24092.25  & 24092.25 & \textbf{3951.26} & \textbf{3951.26} & \textbf{3951.26} \\ \hline
	    \begin{tabular}[c]{@{}c@{}}1/64x\end{tabular} & 7367.46 & 23937.48 & 23924.91 & \textbf{3700.71} & \textbf{3920.50}  & \textbf{3916.37} \\ \hline
	    \begin{tabular}[c]{@{}c@{}}1/32x\end{tabular} & 5466.10 & 23769.53 & 23826.57 & \textbf{3508.57} & \textbf{3916.17} & \textbf{3910.60}  \\ \hline
	    \begin{tabular}[c]{@{}c@{}}1/16x\end{tabular} & 4261.62  & 23290.30 & 23542.19 & \textbf{3200.06} & \textbf{3890.24} & \textbf{3905.22} \\ \hline
	    \begin{tabular}[c]{@{}c@{}}1/8x\end{tabular}  & 3139.47  & 22475.57  & 22837.19 & \textbf{2760.77} & \textbf{3633.68} & \textbf{3872.23} \\ \hline
	    \begin{tabular}[c]{@{}c@{}}1/4x\end{tabular} & \textbf{2249.94}  & 20584.628 & 20962.07 & 2252.38 & \textbf{2614.04} & \textbf{3716.86} \\ \hline
	    \begin{tabular}[c]{@{}c@{}}1/2x\end{tabular} & \textbf{1646.55}  & 5741.87  & 16050.62 & 1822.76 & \textbf{1696.34} & \textbf{3103.81} \\ \hline
	    \begin{tabular}[c]{@{}c@{}}1x(full)\end{tabular} & \textbf{1343.33}  & \textbf{1343.33}   & \textbf{1343.33}  & 1593.37 & 1593.37 & 1593.37 \\ \hline
			\noalign{\vskip 2mm}
	    \end{tabular}%
    }
    }
  \end{minipage}
	\begin{minipage}{1\textwidth}
		\def\arraystretch{2}
		\subfigure[RMSE metric result (mm) in the horizontally sampled scan]{
		\scalebox{0.82}{
		\def\arraystretch{1,0}
		\begin{tabular}{c|c|c|c|c|c|c|c|c} \hline
		Point density& \multicolumn{4}{c|}{Ours} & \multicolumn{4}{c}{Ma et al. \cite{Ma2018}} \\ \cline{2-9}
		(\# scan lines) & RMSE [mm] & MAE [mm] & iRMSE [1/km] & iMAE [1/km] &
									 RMSE [mm] & MAE [mm] & iRMSE [1/km] & iMAE [1/km] \\ \hline
		\begin{tabular}[c]{@{}c@{}}1/4x\\ (16)\end{tabular} &
			\textbf{2614.04} &
			\textbf{1141.02} &
			\textbf{14.89} &
			\textbf{5.68} &
			20584.63 &
			13550.55 &
			54.97 &
			43.45 \\ \hline
		\begin{tabular}[c]{@{}c@{}}1/2x\\ (32)\end{tabular} &
			\textbf{1696.34} &
			\textbf{705.02} &
			\textbf{18.17} &
			\textbf{4.36} &
			5741.87 &
			3132.26 &
			45.32 &
			30.65 \\ \hline
		\begin{tabular}[c]{@{}c@{}}full\\ (64)\end{tabular} &
			1593.37 &
			547.00 &
			27.98 &
			2.36 &
			\textbf{1343.33} &
			\textbf{358.66} &
			\textbf{4.28} &
			\textbf{1.64} \\ \hline
			\noalign{\vskip 2mm}
		\end{tabular}
		}

		}
		\label{tab:kitti_metric}
	\end{minipage}%
  \caption{Quantitative comparison to \citeauthor{Ma2018}~\cite{Ma2018} in the KITTI depth completion dataset regarding sparse and biased range data. Best performance is marked in bold. The point density of 0\% represents the result of monocular depth estimation \cite{Godard2018} without the proposed method.}
  \label{fig:kitti_quan}
	\vspace{-4mm}
\end{figure*}



\subsection{KITTI Depth Completion}

The KITTI depth completion dataset provides RGB images, \ac{LiDAR} raw data, and ground-truth depth data of outdoor scenes. The ground-truth data were generated by accumulating 11 sparse laser, and the outliers of laser scans were removed using stereo reconstruction \cite{Uhrig2017}. The evaluation of \ac{DC} is performed with these semi-dense ground-truth points. This dataset includes 85,898 training data, 1,000 validation data, and 1,000 test data. We evaluated the proposed method and a state-of-the-art method in the validation set.

\figref{fig:kitti_qual} and \figref{fig:kitti_qual2} show the qualitative performance of the proposed method. The proposed method utilizes monodepth2~\cite{Godard2018} as a front-end module. As shown in the blue box in \figref{fig:kitti_qual}(e), a blurry shape of a man riding a bicycle takes the output of MDE. However, the proposed method can be used as a back-end module given sparse depth measurements so that the man's shape becomes sharper and more accurate, as shown in the blue box in \figref{fig:kitti_qual}(c). The proposed method also outperforms a state-of-the-art \ac{DC} method \cite{Ma2018}, as shown in the red boxes in \figref{fig:kitti_qual}(c) and (d). Our module effectively sustains the state-of-the-art depth inference of monodepth2 at the region where the \ac{LiDAR} is not received. \figref{fig:kitti_qual}(f) shows the overall comparison of the proposed method to monodepth2 and Ma et al. \cite{Ma2018}. Furthermore, \figref{fig:kitti_qual2} represents the \ac{DC} results under uniformly and horizontally sampled range data. This shows that Ma et al. \cite{Ma2018} found it difficult to cope with the changes in various range data, whereas our proposed module enhances \ac{MDE} effectively with the given \ac{LiDAR} data.

Next, we present the quantitative results of the proposed methods in terms of sparsity, biased measurements, and modality change.

\begin{figure*}[!t]
	\def\width{1.0\textwidth}%
	\centering
	\includegraphics[width=\width]{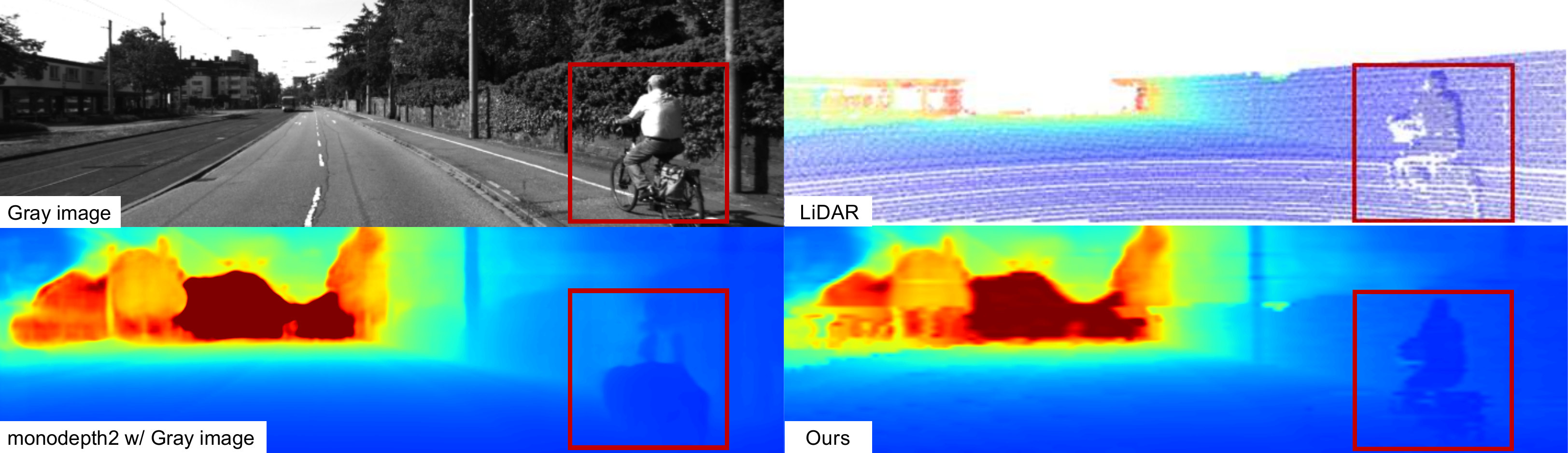}
	\caption{Qualitative result of the proposed method that used a different imaging modality. For a given gray image on the top left, the red box represents a man riding a bicycle. The depth image is worse than its original output (bottom left) due to the different modality. The proposed method successfully recovers the shape of a man riding a bicycle (bottom right).}
	\label{fig:kitti_modality}
	\vspace{-6mm}
\end{figure*}

\subsubsection{Sparsity of \ac{LiDAR} Measurements}

We uniformly sampled the eight-level ratio from the full-scan data (\figref{fig:sampling_methods_KITTI}) to see the effects of sparsity. We compared the proposed method using unsupervised \ac{MDE} \cite{Godard2018} against a self-supervised \ac{DC} method \cite{Ma2018}. We use the open-sourced version of the \citeauthor{Ma2018}~\cite{Ma2018} for this evaluation. As shown in \figref{fig:kitti_quan}(a) and (b), the results reveal that the proposed method presents comparability with one-fourth of the samples of full \ac{LiDAR} scans and that it outperforms after one-eighth of the \ac{LiDAR} scan, whereas \cite{Ma2018} performed better than us by \unit{250.04}{mm} on the RMSE metric when the full \ac{LiDAR} scans are used. As aforementioned, the deep-learning-based methods are able to train the LiDAR point's confidence in a given LiDAR range map, whereas our method only picks the closest point in the resized range map. We consider that this effect deteriorates our results when the number of \ac{LiDAR} points increases. Note that the RMSE error of our method increases until the level of \ac{MDE}, whereas the RMSE error of \citeauthor{Ma2018}~\cite{Ma2018} increases as a power function $cx^p$ as described in their paper. This means that our method performs more stably in sparsity variance, which is especially important in robotics usage cases.

\subsubsection{Biased \ac{LiDAR} Measurements}

The proposed method is not only robust toward data sparsity but also toward the distribution pattern of bias. This may involve installing configuration and various \ac{FOV} specifications of the perceptual sensors. For validation, we sampled the full \ac{LiDAR} data horizontally and vertically to see the effects of biased \ac{FOV}. The KITTI dataset originally provided biased \ac{LiDAR} data in terms of RGB images for the limited \ac{FOV} of a \ac{LiDAR} sensor. As a result, we were unable to evaluate the depth accuracy where the \ac{LiDAR} points could not reach. As alternative methods, we divided the \ac{LiDAR}'s \ac{FOV} horizontally and vertically and used \ac{LiDAR} raw data on only a portion of the area so that we could evaluate the remaining areas that the \ac{LiDAR} points did not exist. As can be seen in \figref{fig:kitti_quan}, the proposed method also outperforms the state-of-the-art method \cite{Ma2018} except for the case of a full \ac{LiDAR} scan. Compared to the uniformly sampled sets, the \ac{RMSE} of \citeauthor{Ma2018}~\cite{Ma2018} increased rapidly. From the experimental results, we consider that separating a depth inference module and its depth regression would be advantageous for the stability of DC against eccentric measurements.

\begin{figure*}[!t]
	\centering
  \begin{minipage}{0.5\textwidth}
  \subfigure[RMSE comparison to \cite{Ma2017} and \cite{Yang2019_DDP}]{
     \includegraphics[trim=5cm 15cm 3cm 9cm clip, width=\textwidth]{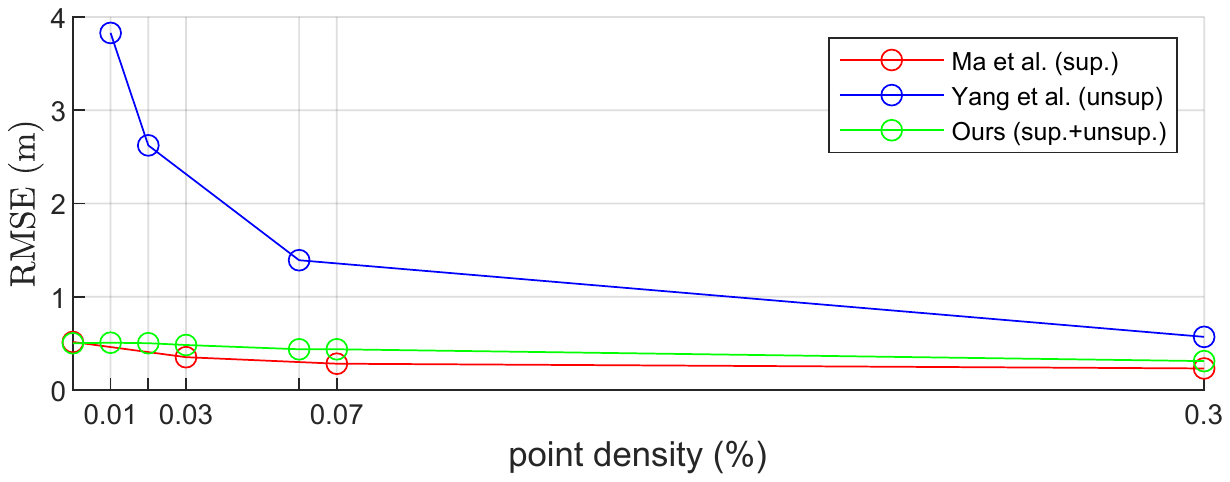} 
	   \label{fig:graph_results_NYU}
  }
  \end{minipage}%
  \begin{minipage}{0.4\textwidth}
  \subfigure[RMSE metric result (mm)]{
  \label{tab:results_NYU}
  \scriptsize
		\begin{tabular}{c|c|c|c}
		\hline
		  \begin{tabular}[c]{@{}c@{}}point density (\%)\\ (\#points)\end{tabular} &
		  \begin{tabular}[c]{@{}c@{}}Ma et al.\cite{Ma2017} \\ (sup.)\end{tabular} &
		  \begin{tabular}[c]{@{}c@{}}Yang et al.\cite{Yang2019_DDP}\\ (unsup.)\end{tabular} &
		  Ours w/ \cite{Yin2019} \\ \hline
		0 (0)      & 514 & -    & 500.31 \\ \hline
		0.01 (7)   & -   & 3829 & 506.3  \\ \hline
		0.02 (14)  & -   & 2623 & 500.92 \\ \hline
		0.03 (20)  & 351 & -    & 481.6  \\ \hline
		0.06 (42)  & -   & 1391 & 435.85 \\ \hline
		0.07 (50)  & 281 &      & 435.85 \\ \hline
		0.3 (200) & 230 & 569  & 308.89 \\ \hline
		\noalign{\vskip 2mm}
		\end{tabular}%
  }
  \end{minipage}
  \caption{Quantitative comparison to \citeauthor{Ma2017}~\cite{Ma2017} and \citeauthor{Yang2019_DDP}~\cite{Yang2019_DDP} in the NYU-Depth-V2 dataset. The point density of 0\% represents the result of monocular depth estimation \cite{Yin2019} without the proposed method.}
	\label{fig:quan_nyu}
	\vspace{-4mm}
\end{figure*}

\subsubsection{Robustness to Modality Variance}
\label{sec:modality}

In \figref{fig:kitti_modality}, we show the qualitative result that the proposed method can be adapted to another front-end \ac{MDE} used with a gray-scaled input image. When \ac{MDE} is designed, it is considered within the properties of its own imaging modality. For example, the \ac{MDE} using thermal \ac{IR} images \cite{Kim2018} discusses the unique features of thermal images, such as blurry imaging and thermal time variance, in its architecture. By exploiting its own best performance on imaging depth inference, the proposed method only focuses on the depth-to-depth regression, in which \ac{GP} can perform strongly in the theoretically proven way.

\subsection{NYU-Depth-V2}
\label{sec:nyu}

The NYU-Depth-V2 dataset provides RGB and depth image, as well as accelerometer data collected by Microsoft Kinect. As a subset of the raw data, the labeled data consist of synchronized pairs of RGB and depth images with dense labels. The dataset is officially split into 795 training images and 654 test images. We evaluated the proposed method in the test set. As sparse range measurements in this experiment, we randomly selected a uniform and assigned number of depth points from the ground-truth depth map.

\begin{table}[!b]
\centering
	\footnotesize
		\begin{tabular}{c|c|c|c|c}
		\hline
		\# points & RMSE & REL & $\delta_1$ & $\delta_2$ \\ \hline
		0         & 500.31        & 10.79        & 87.94             & 97.52             \\ \hline
		50        & 435.85        & 8.89         & 90.77             & 98.18             \\ \hline
		200       & 308.89        & 5.76         & 95.22             & 99.06             \\ \hline
		500       & 230.66        & 3.97         & 97.40             & 99.54             \\ \hline
		1000      & 186.85        & 3.13         & 98.40             & 99.77             \\ \hline
		\end{tabular}%
	\caption{\small Overall results with VNL\cite{Yin2019} in the NYU-Depth-V2}
	\label{tab:metric_nyu}
\end{table}

\begin{figure}[!t]
	\def\width{1\columnwidth}%
	\centering
	\includegraphics[width=0.9\width]{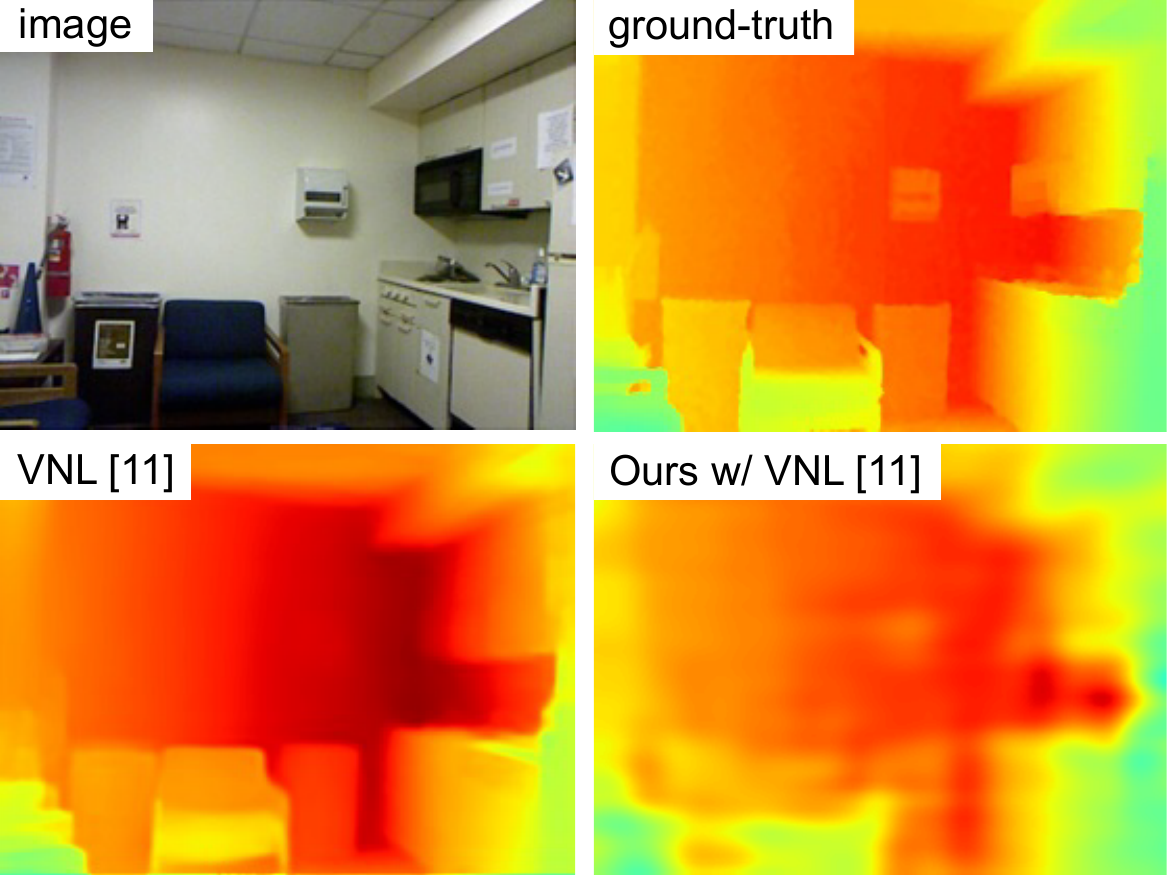}
	\caption{Qualitative result in the NYU-Depth-V2 dataset. Our method successfully recovers an indoor depth image given 500 uniformly distributed samples (point density of 0.7\%).}
	\label{fig:qual_nyu}
\end{figure}

Since we could not find an open-sourced and unsupervised \ac{MDE} method implemented in NYU-Depth-V2, we inevitably used a supervised \ac{MDE} method, VNL~\cite{Yin2019} as our front-end module. Therefore, we compared our results with a supervised method \cite{Ma2017} and an unsupervised method \cite{Yang2019_DDP}, as illustrated in \figref{fig:quan_nyu}. The proposed method outperforms the state-of-the-art unsupervised method \cite{Yang2019_DDP} in every point density case, and it is also comparable with a recent supervised method \cite{Ma2017}. Note that the proposed methods may have the potential for improvement considering the ability to adapt with state-of-the-art \ac{MDE}, although the RMSE of \citeauthor{Ma2017}~\cite{Ma2017} is \unit{79}{mm} higher in the case of 200 samples' case (\tabref{tab:metric_nyu}). In the overall results of our method, we experienced that the depth map of MDE becomes more blurred after passing our method. We consider this is due to the fact that \ac{GP} does not directly understand object shape. Therefore, we remain that our method can be improved with semantic segmentation or image-guided edge-preserving methods for future work. The quantitative and qualitative results are shown in \tabref{tab:metric_nyu}, \figref{fig:quan_nyu}, and \figref{fig:qual_nyu}.


\section{Conclusion}
\label{sec:conclusion}

In this paper, we introduced a depth-regression method for \ac{DC} using sparse-but-accurate depth measurements and dense-but-inaccurate depth inference from deep-learning-based estimation. Unlike the other \ac{DC} methods, we divided the \ac{DC} architecture into two steps (i.e., \ac{MDE} and depth regression using \ac{GP}), so that we could achieve robust and accurate performance even when given sparse and biased range measurements. We demonstrated that the proposed method outperformed state-of-the-art unsupervised methods in KITTI depth completion and NYU-Depth-V2 datasets with regard to the sparsity and bias of measurements. We also showed our method's flexibility, which can be used as a plugin module in place of other depth-estimation modules that may have a different imaging modality. We believe that the proposed methods can be used in a SLAM application that needs a dense visualized map and that only observes a few biased feature points in its environment.

\small
\balance
\bibliographystyle{IEEEtranN} 
\bibliography{string-short,root}

\end{document}